\documentclass[conference]{IEEEtran}
\IEEEoverridecommandlockouts
\usepackage{cite}
\usepackage{amsmath,amssymb,amsfonts}
\usepackage{algorithm}
\usepackage[noend]{algpseudocode}
\usepackage{graphicx}
\usepackage{textcomp}
\usepackage{xcolor}
\usepackage{lipsum}
\usepackage{flexisym}
\usepackage{amsfonts,amssymb}
\usepackage{array}
\usepackage{multirow}

\def\BibTeX{{\rm B\kern-.05em{\sc i\kern-.025em b}\kern-.08em
    T\kern-.1667em\lower.7ex\hbox{E}\kern-.125emX}}
    
\begin{document}

\title{Unlucky Explorer: A Complete non-Overlapping Map Exploration\\
}


\author{Mohammad Sina Kiarostami$^{\ast}$, Saleh Khalaj Monfared$^{\ast}$, Mohammadreza Daneshvaramoli$^{ \ast}$, Ali Oliayi$^{\dagger}$, \\ Negar Yousefian$^{\dagger}$, Dara Rahmati$^{ \ast}$, 
 Saeid Gorgin$^{\ast}$
\\
$^{\ast}$School of Computer Sciences, Institute for Research in Fundamental Sciences (IPM), Tehran, Iran\\
$^{\dagger}$ K.N.Toosi University of Technology, Tehran, Iran
\\
\{skiarostami, monfared, daneshvaramoli, dara.rahmati, gorgin\}@ipm.ir\\ \{ali.oliayi, negaryousefian\}@email.kntu.ac.ir 
}

\maketitle

\begin{abstract}

Nowadays, the field of Artificial Intelligence in Computer Games (\textit{AI in Games}) is going to be more alluring since computer games challenge many aspects of AI with a wide range of problems, particularly general problems. One of these kinds of problems is \textit{Exploration}, which states that an unknown environment must be explored by one or several agents.
\par In this work, we have first introduced the \textit{Maze Dash} puzzle as an exploration problem where the agent must find the a \textit{Hamiltonian Path} visiting all the cells. Then, we have investigated to find suitable methods by a focus on \textit{Monte-Carlo Tree Search} (MCTS) and \textit{SAT} to solve this puzzle quickly and accurately. An optimization has been applied to the proposed MCTS algorithm to obtain a promising result. Also, since the prefabricated test cases of this puzzle are not large enough to assay the proposed method, we have proposed and employed a technique to generate solvable test cases to evaluate the  approaches. Eventually, the MCTS-based method has been assessed by the auto-generated test cases and compared with our implemented \textit{SAT} approach that is considered a good rival. Our comparison indicates that the MCTS-based approach is an up-and-coming method which could cope with the test cases with small and medium sizes with faster run-time compared to SAT. However, for certain discussed reasons, including the features of the problem, tree search organization, and also the approach of MCTS in the \textit{Simulation} step, MCTS takes more time to execute in Large size scenarios. Consequently, we have found the bottleneck for the MCTS-based method in significant test cases that could be improved in two real-world problems.

\end{abstract}


\section{Introduction and Background}
\textit{Graph Traversal} is an important and famous problem in computer science with many applications in memory and storage systems \cite{aggarwal1988input}, network flow \cite{cheung1983graph}, as well as computer games \cite{plaat1996exploiting}.
Advances in different aspects such as insuring the security \cite{karvandi2020way}, efficient path fining \cite{bobrow1985time}, and \textit{Graph Exploration} in the robotics are also addressed by the researchers \cite{fraigniaud2005graph}. As the conventional approaches to solving this problem and other variations like \textit{Tree Traversal}, Depth-first search (DFS), and Breadth-first search (BFS) are known to be effective in general. On the other hand, random-based approaches such as Monte-Carlo Tree Search (MCTS) are demonstrated to be efficient in many search-based problems and games as well as \cite{6145622}. 

The fundamental problem of many simple computer games lays on solving specific computer or mathematical puzzles. The solution methodology used in many of these games is very relevant to fundamental approaches. For instance, \textit{Flow-Free} is a variant of a known mathematical puzzle named \textit{Numberlink}, and interestingly, the problem could be addressed as a Multi-Agent Path Finding \cite{8848043}. In this context, \textit{Icosian Game} \cite{pegg2009icosian} as an old mathematical game invented by W.R Hamilton could be considered as a modified version of a Graph Traversal problem. The objective in Icosian is finding a \textit{Hamiltonian cycle} along the edges of a dodecahedron, visiting all the vertexes of the graph by ending at the same point as the starting vertex. The Hamiltonian Path problem as an NP-Complete problem \cite{hartmanis1982computers} has its own applications in various fields \cite{cooper2019hamiltonian} with many solution methods \cite{bjorklund2014determinant}.

In this article, we investigate the foundation of the Maze Dash game. We demonstrate that the constraints involved in solving the game, inevitably minimize the number of \textit{Turning Movement} in the grid exploration procedure. Satisfying this particular condition applied in this game could be interesting in terms of real-world robot exploration since an extra cost is often associated with the \textit{Turning Movement} in smart explorer vehicles \cite{davoodi2015clear}. Hence, finding an efficient and effective solution to the focused Maze Dash game could lead to faster \textit{Grid Traversal} approaches where realistic restrictions in the robots are considered. 

Furthermore, We mathematically define the underlying primary problem of the game as a particular case of a Hamiltonian Path problem. Then, by studying the problem's specifications, we tackle the problem with different possible approaches, including the MCTS, as one of the promising methods. We examine the unique characteristics of the involved tree search in detail and study the exclusive attributes of Hamiltonian Path in this problem.

\subsection{Maze Dash Game}
 Maze Dash is a puzzle game with a single agent and a 2-Dimensional grid map. The map might have some obstacles or blocking cells. The agent moves in the map and marks the cells after visiting them by changing their color and can not return to the marked cells. So, each cell must be visited just once. Eventually, the puzzle aims to visit all the cells or to explore the whole of the map.
\par
\begin{figure}[t]
  \includegraphics[width=0.4\textwidth]{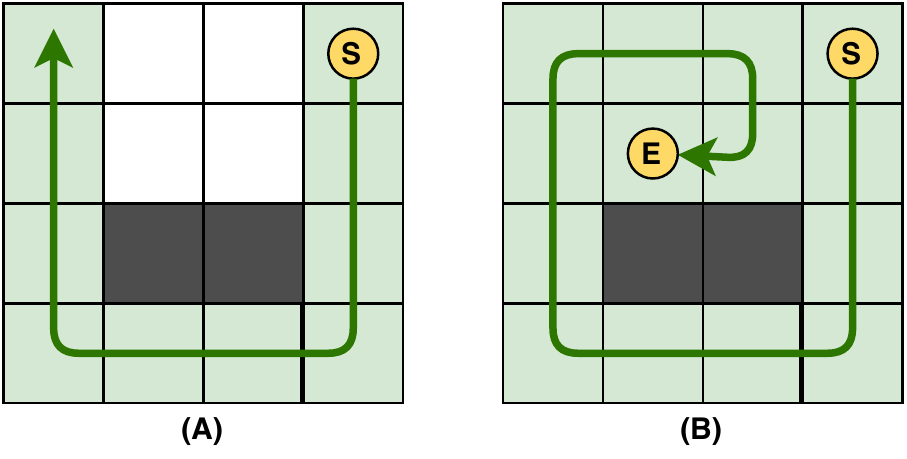}
  \centering
\caption{\textit{Maze Dash} game solving process} 
\label{maze}
\end{figure}

 The essential rule in this puzzle is that if the agent chooses to go to the one the quad directions, it will continue to move until it reaches an obstacle or wall. As shown in Figure \ref{maze}, the agent starts to move from the initial cell (S) and decides to go down to reach the wall or the border of the grid. Then, it keeps moving to explore all of the possible cells, finishing the traversal at the last cell (E).

\subsection{Monte-Carlo Tree Search}
MCTS is a best-first search algorithm with four main steps which are \textit{Selection}, \textit{Expansion}, \textit{Simulation}, and \textit{Backpropagation}. This algorithm uses Monte-Carlo methods to sample steps and create the search tree indeterminately to solve problems in their particular domain \cite{6145622}. Like other tree-based approaches, the algorithm requires considering an initial state as its root to construct the search tree. In the context of our problem, the initial state is the state that the agent craves to move from its current cell. 
\par The first step of the MCTS is the selection that the algorithm chooses the best node, which is a leaf at the moment based on the \textit{Tree Policy}. Then, at the expansion point, all non-terminal children of the selected node, if exist, will be expanded. In the next step, simulation, MCTS strides in the search tree aimlessly based on a policy until it reaches a leaf. The obtained result will be evaluated and measured how much is this result is analogous to the desired result, and how many of the rules and conditions of the problem are satisfied. Finally, in the backpropagation, the results are propagated back through the tree, and all related node values are updated. After that, the next rounds will be iterated to find the suitable solution.

\par

\section{Proposed Method}

\subsection{Problem Definition}
The exact definition of the problem as a modification of a \textit{Hamiltonian Path} in a 2-D grid is described below. The the set of $O=\{o_1,o_2,...,o_m\}$ is demonstrated as the \textit{obstacle} set which determines the coordination of the obstacle cells in the grid. By considering an $N\times N$ grid, the function $\pi$ identifies the movement path in the grid:
\begin{equation}  \label{eq1}
\begin{split}
    &\pi : \mathcal{N} \rightarrow \mathcal{N}\times \mathcal{N}
\end{split}
\end{equation}
The input of $\pi$ function rises incrementally to represents the movement path in the grid. The output represents the coordinates with the constrain of moving a single cell at each step to ensure the consistency of the solution path:
\begin{equation} \label{eq:2}
\begin{split}
 &\pi(0)=S\\
 &Direction=\{(1,0),(0,1),(-1,0),(0,-1)\} \\
 &\forall (N^2 - |O|) > i > 0 :\\
 &\pi(i+1)-\pi(i) \in Direction, \pi(i+1) \notin O\\
 &\forall i,j: \pi(i) \neq \pi(j)\\ 
\end{split}
\end{equation}
In Equation \ref{eq:2}, $S$ is the initial coordinate of the beginning cell. The Direction set $D$ is the set of all possible movements that can be used in this context. As for the constraints regarding the minimum turning movement restrictions in the game, the $\pi$ function falls into either one of the two \textit{Straight Movement}, \textit{Turning  Movement} conditions as defined in Equation \ref{eq:3}, respectively:
\begin{equation} \label{eq:3}
\begin{split}
 &\forall (N^2 - |O|) > i > 1 : one of Three: \\
 &Straight:\pi(i) + (\pi(i)- \pi(i-1)) = \pi(i+1)\\
 &Turning:\pi(i) + (\pi(i)- \pi(i-1))\in O \\
 &Turning:\exists j < i:\pi(i) + (\pi(i)- \pi(i-1)) = \pi(j) \\
\end{split}
\end{equation}
Note that the turning is occurred either by a blocking obstacle or a previously occupied cell by the earlier path. 
Hence, the $i^{th}$ step in a cell must be as the same previous direction, or a turning movement happens.

\par Eventually, the solution of the game comprise of the adequate assignment for the $\pi$ function satisfying all the constraints presented in  Equations \ref{eq:2} and \ref{eq:3}.

\subsection{Promising Approaches}

\subsubsection{SAT}
The accurate mathematical definition discussed, as the foundation of the problem, could be fed to a Satisfiability solver (SAT Solver, e.g., \textit{Z3}\cite{de2008z3}) as a simple solution approach by converting the conditions into boolean constraints. In this respect, the boolean assignments of $t,s : \mathcal{N} \rightarrow \{0,1\}$, should be defined to determined whatever a cell falls into the \textit{Turning Movement} or the \textit{Straight Movement} sets. Then, by creating the same constraints in the function $\pi$ for each cell of the grid, the Satisfiability check could be performed over the described assignments of $\{t,s,\pi\}$.

\begin{figure}[h!]
  \includegraphics[width=0.3\textwidth]{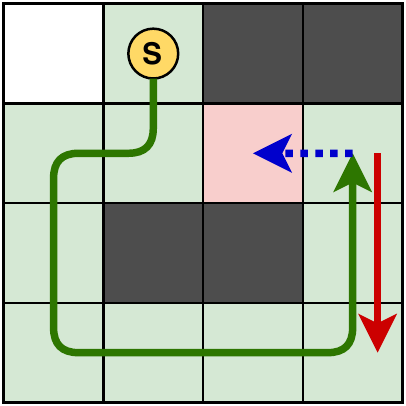}
  \centering
\caption{Procedure of Backtracking in solving the puzzle} 
\label{back}
\end{figure}

\subsubsection{Backtracking}

As a naive approach to solve the problem, one could apply the \textit{Backtracking} method to find the correct solution. The backtracking process is very similar to a \textit{DFS} method. Considering the conditions of the game, a single branch of the possible solution is pursued until all the cells are traversed or a deadlock occurs. In the case of a failure, the movement path backtraces itself to the previous state and changes the branch by choosing another possible path. A simple backtrack demonstration is shown in Figure \ref{back}. The algorithm has to eliminate its current path to correct it since one of the cells is not visited.

\subsubsection{MCTS}
\begin{figure}[!h]
  \includegraphics[width=0.3\textwidth]{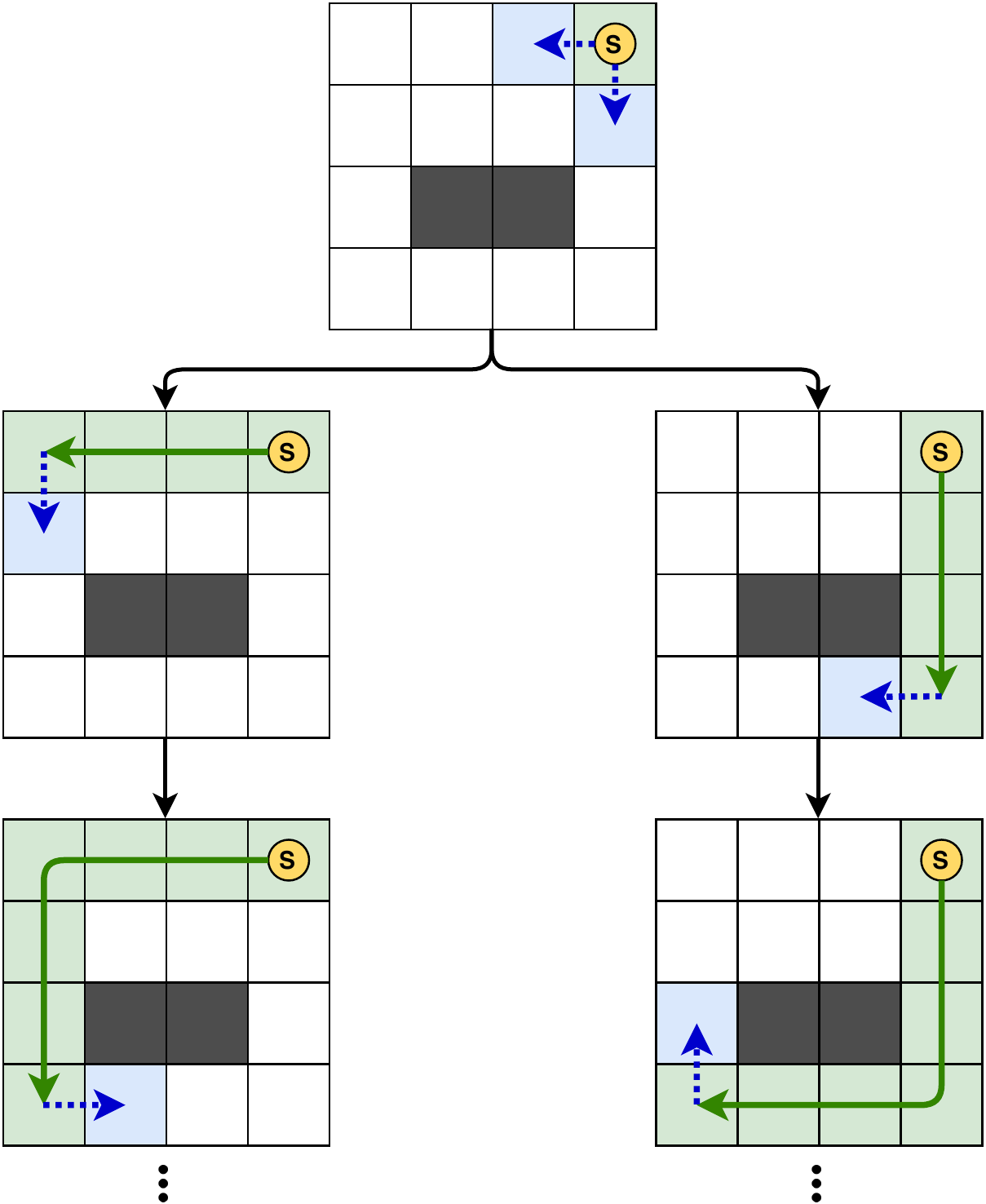}
  \centering
\caption{Construction of the tree search of the puzzle} 
\label{tree}
\end{figure}

As mentioned in the background section, one of the particular and pronounced features of the game is that the agent must explore the grid with the minimum number of turns. More precisely, the agent only could change its direction when it reaches the end of the current path. This constraint intensely affects the tree search of the puzzle. As shown in Figure \ref{tree}, each non-terminal node of the tree search has only one or two children. Therefore, the \textit{Branch Factor} of the tree is equal to or smaller than two making the tree significantly long in-depth.
One of the approaches that could solve the problem is MCTS. As explained earlier, MCTS runs its four steps iteratively to construct the tree search and find the solution.

To solve the problem more efficiently, we have disregarded the nodes with a certain failure at their final stats to prune the tree. More precisely, the Expansion function was amended to block a terminal state with an inadequate value of the desired final state. Then, the Selection function blocks a node if all of its children were blocked before, by assigning 0 as the value of the node. Therefore, gradually wrong paths could be blocked efficiently. Nevertheless, even by employing these modifications to the algorithm, the execution time increases immensely due to the computational overhead caused by huge depth in the tree. Another optimization has been employed by applying the \textit{Fast Rollout Function}, where instead of constructing a new state for each \textit{Child Node}, the state of the parent is updated each time. This optimization reduces the memory consumption of $\mathcal{O}(N^2)$ to $\mathcal{O}(1)$ for each node \cite{daneshvaramoli2019decentralized}. These details will be discussed in the result section.

Note that, another approach to the problem could be a BFS method where all of the branches are searched at the time without traversing deep into the final state. Not surprisingly, this method burdens considerable memory usage to store the branch states. This memory overuse could be predicted due to the tremendous depth of the tree search. Consequently, a BFS approach would not be a suitable solution compared to other possible methods.

\section{Evaluation}

As the test cases of the Maze Dash game are not large enough to assess the methods correctly, we decided to utilize a method to generate large random test cases. First, an empty grid with the desired size is assumed. Then an agent starts to move through the grid randomly. Whenever the agent turns or changes its direction, an obstacle is placed at the next cell of the current cell. It means that there was a hypothetical obstacle in the path, so the agent decided to change the direction. For avoiding creating a unique path for solving the test case, we defined a variable as the number of obstacles. After generating the test cases, all of them have been tested by the Backtracking approach to make sure that they are solvable. 

All of the compiled files were executed on a machine running Ubuntu 16.04 equipped with two Intel XEON E5 2697V3 CPUs clocked at 2.6 GHz and 128 GB of DDR3 RAM.


\begin{table*}[h!]
\centering
\Huge \caption{ evaluation of different algorithms for solving \textit{Maze Dash} Game}
\label{tab:1}
\resizebox{1.6\columnwidth}{!}{%
\begin{tabular}{|c|c|c|c|c|c|c|c|}
\hline
\multirow{2}{*}{Grid Size($N^2$)} & \multirow{2}{*}{No Obstacle} & \multicolumn{2}{c|}{SAT} & \multicolumn{2}{c|}{Backtrack(DFS)} & \multicolumn{2}{c|}{Randomize(MCTS)} \\ \cline{3-8} 
                               &                              & Run-Time(S) & Memory(MB) & Run-Time(S)       & Memory(MB)      & Run-Time(S)       & Memory(MB)       \\ \hline
5x5                            & 4                            & 0.45        & 6.01       & 0.010             & 7.7             & 0.011             & 5.94             \\ \hline
6x6                            & 10                           & 0.65        & 8.12       & 0.012             & 8.1             & 0.013             & 10.61            \\ \hline
10x10                          & 32                           & 1.79        & 12.83      & 0.45              & 10.4            & 0.07              & 11.5             \\ \hline
15x15                          & 66                           & 4.18        & 13.27      & 5.24              & 13.5            & 0.09              & 14.2             \\ \hline
20x20                          & 133                          & 11.28       & 13.86      & 9.62              & 19.2            & 1.16              & 16.9             \\ \hline
30x30                          & 378                          & 17.94       & 15.5       & 43.7              & 26.1            & 5.42              & 20.1             \\ \hline
50x50                          & 776                          & 126.6       & 49.9       & Failed            & Failed          & Failed            & Failed           \\ \hline
\end{tabular}%
}
\end{table*}

As shown in Table \ref{tab:1}, we have implemented and compared possible approaches to solve the defined problem. Each test case is executed by each method, 50 times, and the average values are presented. It is worth mentioning that the \textit{Backtracking} algorithm is used to indicate our worst-case scenario, not to be a good rival. The MCTS approach performs well in small and medium-size test cases but could not cope with large ones. The results of the MCTS method could be discussed. First, most of the execution run-time of the algorithm is spent in the simulation step. Assume that $ I$ defines the number of iterations of the simulation in each MCTS traversal. In each simulation, the algorithm would traverse the tree down to $N^{2}$ depth. After selecting each node and adding one depth to the MCTS tree, the algorithm would be repeated. Thus, the search would be performed for another $N^{2}$ times. Ultimately, simulations process enlarges and will have an immense cost, calculated as follows:

\begin{equation} \label{eq:6}
Simulations Cost = I \times N^{4}\\
\end{equation}

This order is a huge cost for the problem since the agent could not determine or predict a complete solution until it tests all possible movements. For instance, the \textit{Evaluation Function} could return 0.98 as a value of a final state, meaning that only 2 cells are not visited among 100. We know that this state is not the accurate answer but the algorithm would recognize it as a promising state in the previous steps. So, the algorithm would never prune these kinds of states. Furthermore, owing to the non-overlapping feature of this modification of the exploration, after each movement, the agent would be faced by new obstacles. Moreover, previously visited cells are considered as dynamic obstacles and walls, which makes the algorithm unable to be optimized by any pre-process methods. Furthermore, as explained, the branch factor of the problem is equal or smaller than 2, constructing a tree with a huge depth without many branches. This kind of tree produces an arduous circumstance for MCTS in its simulation step.


\section{Discussion and Conclusion}

In this article, we first introduced the Maze Dash puzzle as a modification of exploration problem or, more precisely, a non-overlapping exploration that could be solved by Hamiltonian Path. This means that the primary purpose of this problem is to visit all of the cells in 2-D grid. We have investigated promising approaches to find a proper solution. Although four methods implemented to compare with each other, the main focus was on MCTS and SAT. As expected, our implemented SAT could solve the auto-generated test cases accurately. Nevertheless, by reducing the number of obstacles in the grid, the execution time increased exponentially.
\par In small and medium-size test cases, MCTS could outperform SAT. In the large test cases, as explored, the simulation time increased uncontrollably since the algorithm recognizes the failure early in the simulation. As our observation, due to the non-overlapping feature of the problem, the agent considers its previous path as dynamic walls or obstacles. Therefore, the algorithm could not be optimized by pre-processing methods, but the SAT performs more beneficial in these test cases.
\par Investigating the introduced intricacy, the authors of this article came to realize two more practical and pronounced problems that should be considered. The first one is reasonably analogous to the current problem but differs in the non-overlapping constraint. It means that the aim of the problem is that an agent must explore all of the environment fast and accurately. However, the agent's previous path would not be defined as a new obstacle. Thus, the agent prefers not to use the visited cells but is not forced to do this.
The second problem is the exploration of a grid to find a goal with minimum numbers of turns, by assuming that turning movement has an additional cost, since the agent must reduce its velocity, stop, and then start to move again \cite{youtube,davoodi2015clear}. By this constraint, the agent prefers to choose a path with lesser turnings. In our future studies, we would concentrate on these two problems, which would be useful in the real world applications, such as 2-D Robotic soccer.


\vspace*{-2mm}


\vspace{2pt}

\bibliographystyle{IEEEtran}
\bibliography{CoG-MazeDash}

\end{document}